\documentclass{article}
\usepackage{PRIMEarxiv}

\usepackage[utf8]{inputenc} 
\usepackage[T1]{fontenc}    
\usepackage{hyperref}       
\usepackage{url}            
\usepackage{booktabs}       
\usepackage{amsfonts}       
\usepackage{nicefrac}       
\usepackage{microtype}      
\usepackage{lipsum}
\usepackage{fancyhdr}       
\usepackage{graphicx}       
\graphicspath{{media/}}     
\usepackage{multirow}
\usepackage{graphicx}
\usepackage{colortbl}
\usepackage{amsmath,amssymb,amsfonts}
\usepackage{textcomp}
\usepackage{xcolor}
\usepackage[symbol]{footmisc}
\usepackage{authblk} 
\hypersetup{
  colorlinks=True,
  citecolor=cyan,
  linkcolor=cyan,
  urlcolor=cyan}

\title{Balancing Exploration and Exploitation: \\Disentangled $\beta$-CVAE in De Novo Drug Design}

\begin{document}
\author[a,$*$]{Guang Jun Nicholas Ang}
\author[a]{De Tao Irwin Chin}
\author[a,b]{Bingquan Shen}

\affil[a]{Engineering Science Programme, National University of Singapore, 117575, Singapore}
\affil[b]{DSO National Laboratories, Singapore}
\affil[$*$]{Corresponding author. Email addresses: anggjnicholas@u.nus.edu}

\maketitle

\begin{abstract}
Deep generative models have recently emerged as a promising \textit{de novo} drug design method. In this respect, deep generative conditional variational autoencoder (CVAE) models are a powerful approach for generating novel molecules with desired drug-like properties. However, molecular graph-based models with disentanglement and multivariate explicit latent conditioning have not been fully elucidated. To address this, we proposed a molecular-graph $\beta$-CVAE model for \textit{de novo} drug design. Here, we empirically tuned the value of disentanglement and assessed its ability to generate molecules with optimised univariate- or-multivariate properties. In particular, we optimised the octanol-water partition coefficient (ClogP), molar refractivity (CMR), quantitative estimate of drug-likeness (QED), and synthetic accessibility score (SAS). Results suggest that a lower $\beta$ value increases the uniqueness of generated molecules (exploration). Univariate optimisation results showed our model generated molecular property averages of ClogP = 41.07\% $\pm$ 0.01\% and CMR 66.76\% $\pm$ 0.01\% by the Ghose filter. Multivariate property optimisation results showed that our model generated an average of 30.07\% $\pm$ 0.01\% molecules for both desired properties. Furthermore, our model improved the QED and SAS (exploitation) of molecules generated. Together, these results suggest that the $\beta$-CVAE could balance exploration and exploitation through disentanglement and is a promising model for \textit{de novo} drug design, thus providing a basis for future studies.
\end{abstract}


\section{Introduction}

Drug discovery aims to identify novel chemical entities with favourable pharmacological properties. However, this is an inherently difficult task due to the immense size and complexity of the chemical search space \cite{polishchuk2013estimation, kirkpatrick2004chemical}. As a result, drug development is a time-consuming and expensive process. Hence, \textit{de novo} drug design is employed to design drug-like compounds based on desired properties in the functional space without requiring a starting template \cite{devi2015evolutionary}. Deep generative models such as conditional variational autoencoders (CVAE)s have shown great success in generating molecules with desired properties. Despite this, CVAEs may face challenges in effectively separating latent molecular representations, leading to a posterior collapse similar to a VAE \cite{richards2022conditional, higgins2017beta, lee2022mgcvae}. To address this issue, we will introduce disentanglement to the CVAE. In brief, disentanglement refers to separating the latent variables in the CVAE’s latent space into independent dimensions that correspond to specific aspects of the data. This paper introduces disentanglement with explicit latent conditioning using a molecular graph-based representation. We propose a $\beta$-CVAE model and introduce multivariate explicit latent conditioning with ClogP, CMR, QED, and SAS as desired molecular properties. We aim to conduct an empirical study on the value of the $\beta$ hyperparameter and present our results.

\section{Background}
\subsection{Representation methods} 

\begin{figure*}
    \centering
    \includegraphics[width=0.8\textwidth]{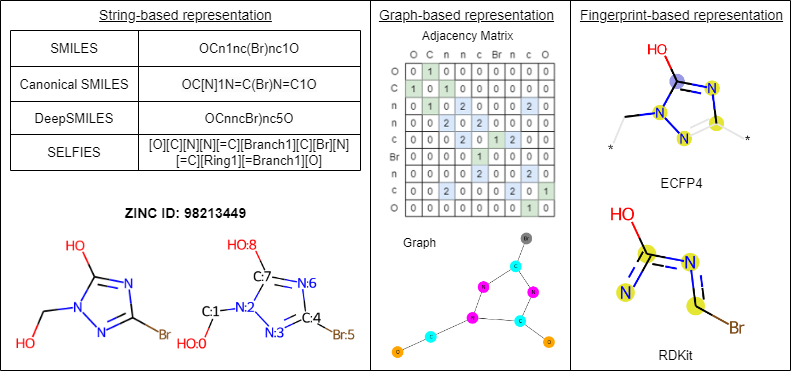}
    \caption{Left: string-based representations. Centre: graph-based representations. Right: Fingerprint-based representations.}
    \label{fig:representation}
\end{figure*}

There are three types of representation methods presented in this study. Figure \ref{fig:representation} depicts an example of the three types of molecular representations of ZINC ID: 98213449. Representation methods allow generative chemists to perform large-scale virtual screening, uncover patterns in molecular structures, and enhance the accuracy of molecular property predictions \cite{weininger1988smiles,carracedo2021review}. 

\subsubsection{String-based representation}
String-based representation is the simplified molecular-input line-entry system (SMILES). SMILES is based on the connectivity of atoms in a molecule as strings of characters \cite{weininger1988smiles, shorten2021text, meyers2021novo}. By using string-based SMILES, generative chemists used recurrent neural networks (RNNs) with long short-term memory (LSTM) to generate molecules via tokenisation \cite{gupta2018generative}. Researchers have recently developed SMILES-based autoencoders (AEs) and variational autoencoders (VAEs) to generate molecules. Although SMILES-based models have been successful, there are some limitations to using SMILES. Firstly, SMILES can sometimes generate multiple representations for the same molecular structure. For example, different start positions and paths through a molecule can produce different SMILES, resulting in a lack of uniqueness in its description. \cite{ lee2022mgcvae, li2018multi}. Secondly, SMILES do not encode all molecular features crucial in biological activity, such as partial charge information and stereochemistry \cite{sousa2021generative}. Lastly, string-based generative models must learn rules unrelated to molecular structure, such as SMILES grammar and atomic order, which can be burdensome for the training process \cite{meyers2021novo}. To overcome these limitations, researchers have developed alternative representations for molecules, including Canonical SMILES \cite{o2012towards}, DeepSMILES \cite{o2018deepsmiles}, and SELFIES \cite{krenn2020self}. These methods offer improved accuracy and efficiency in the representation of molecular structures. Nevertheless, SMILES remains a popular form of representation as large public repositories store molecules in SMILES format \cite{irwin2005zinc,sterling2015zinc}. 

\subsubsection{Graph-based representation}
 
To address the limitations of SMILES, researchers have proposed using graph-based representations for molecules. In this representation, atoms and bonds are nodes and edges, forming a graph that can be represented as a network or an adjacency matrix \cite{meyers2021novo, li2018multi,sousa2021generative}. This approach captures the relationships between constituent atoms and bonds. In addition, graph-based representation can overcome the artificial aspects of SMILES syntax and make it easier to express essential chemical properties such as molecule validity \cite{jin2018junction}. Moreover, this paradigm allows using other forms of deep learning (DL) models, such as graph neural networks (GNNs), for molecule generation.

\subsubsection{Fingerprint-based representation}
Fingerprint-based representation is a molecular descriptor representing a molecular structure in a fixed-length vector of bits, which captures the molecular substructure information or molecular fragments in detail \cite{wen2022fingerprints}. They are commonly used for virtual screening, Tanimoto similarity searches, and evaluating the novelty and uniqueness of generated molecules \cite{segler2018rnn}. An example of a popular fingerprint-based representation is the extended-connectivity fingerprint (ECFP) based on the Morgan algorithm to compact and computationally represent molecular structures \cite{fingerprint2015}.

\subsection{Molecular properties}

Based on the Ghose filter, Crippen's logP (ClogP) and Crippen's molar refractivity (CMR) evaluate potential drug candidates using Crippen's algorithm \cite{ghose1999knowledge, wildman1999prediction}. In brief, ClogP predicts the bioavailability and solubility of a compound, while CMR predicts physical properties such as boiling point and surface tension. The Ghose filter comprises two main chemical properties: octanol-water partition coefficient (logP) and molar refractivity (MR). LogP is defined as the octanol-water partition coefficient of a molecule between aqueous and lipophilic phases and is directly related to lipophilicity. It is also included in Lipinski's rule of five \cite{lipinski2004lead}. MR, on the other hand, is a measure of the total polarisability of a mole of a substance. Filtering unlikely drug-like molecules using the Ghose filter reduces the number of compounds that must be screened, making the drug discovery process more efficient.

Other optimisation properties include the quantitative estimate of drug-likeness (QED) and synthetic accessibility score (SAS). QED is a measure that reflects the likelihood that a compound is a successful drug candidate \cite{craig1998molecular}. SAS measures the ease of synthesising a compound based on its molecular structure \cite{ertl2009estimation}. Both QED and SAS are often used to evaluate further or optimise molecules that primarily satisfy the Ghose filter.

\subsection{Related Work}
\subsubsection{String-based generative models}
Earlier works in deep generative modelling for \textit{de novo} drug design have utilised string-based notation to represent molecules as sequences of tokens. Since SMILES notation is read sequentially from left to right, sequential DL models such as RNNs are well-suited for string-based molecular generation tasks. Segler et al. \cite{segler2018rnn} trained RNNs as generative models for molecular structures, similar to statistical language models in NLP. Similarly, Bjerrum et al. \cite{2017rnn} successfully implemented RNNs with LSTM cells to generate novel molecules. Grisoni et al. \cite{BIMODAL} addressed the limitations of the unidirectional characteristic of RNNs by introducing a bidirectional generative RNN-based model for SMILES-based molecule design, known as BIMODAL. Likewise, REINVENT2.0 is a stacked RNN that uses reinforcement learning (RL) techniques to generate novel molecules from SMILES \cite{reinvent2}. Guimaraes et al. \cite{ORGAN} developed ORGAN, a generative adversarial network (GAN), in a RL framework to propose SMILES with optimised molecular properties.

\begin{table}[t!]
\centering
\caption{Summary of related work.}

\begin{tabular}{llll}
\toprule[1pt]
Model & Representation method & Architecture & Dataset \\ \midrule[0.5pt]
Bjerrum \textit{et al.} \cite{2017rnn} & SMILES & RNN + LSTM & ZINC \cite{sterling2015zinc} \\
Segler\textit{et al.} \cite{segler2018rnn} & SMILES & RNN & ChEMBL \cite{chembl}\\
Gómez–Bombarelli \textit{et al.} \cite{gomez2018automatic} & SMILES & VAE & ChEMBL, QM9 \cite{chembl,qm9}\\
$\beta$-CVAE \cite{richards2022conditional} & SMILES & $\beta$-CVAE & ZINC\cite{sterling2015zinc} \\
REINVENT 2.0 \cite{reinvent2} & SMILES & RNN + RL & ChEMBL \cite{chembl}\\
ORGAN \cite{ORGAN} & SMILES & GAN + RL & ZINC \cite{sterling2015zinc}\\
BIMODAL \cite{BIMODAL} & SMILES & RNN & ChEMBL22 \cite{chembl}\\
GraphVAE \cite{simonovsky2018graphvae} & Graph (Atom based) & GCN + VAE & ZINC\cite{sterling2015zinc} \\
MolecularRNN \cite{molecularrnn} & Graph (Atom-based) & RNN + RL & ChEMBL \cite{chembl}\\
MGCVAE \cite{lee2022mgcvae} & Graph (Atom-based) & CVAE & ZINC \cite{sterling2015zinc}\\
MolGAN \cite{de2018molgan} & Graph (Atom-based) & GAN + RL & ChEMBL\cite{chembl} \\
JT-VAE \cite{jin2018junction} & Graph (Fragment-based) & VAE & ZINC \cite{sterling2015zinc}\\ 
GCPN \cite{gcpn} & Graph (Fragment-based) & GCN + RL & ZINC\cite{sterling2015zinc} \\ \bottomrule[1pt]
\end{tabular}%
 \label{tab: summary of related work}
\end{table}

\subsubsection{Graph-based generative models} 
\textbf{Atom based.} Previous research on graph generation can be categorised into two main types: atom-based and fragment-based models \cite{meyers2021novo}. In atom-based models, SMILES are generated before translating them into graphs via deterministic mappings using RDKit \cite{landrum2013rdkit}. In a similar concept to RNNs with RL, Popova et al. \cite{molecularrnn} introduced MolecularRNN to generate diverse graphs with optimised drug-like properties. Using graph-based representation, Blaschke et al. \cite{reinvent2} enhanced the REINVENT2.0 framework. Deviating from sequence-based generation, GraphVAE and MolGAN learn to generate graph adjacency in a one-shot fashion. GraphVAE uses graph convolutional layers within the VAE framework to output a probabilistic fully-connected graph and uses a graph matching algorithm to compare it to the ground truth \cite{simonovsky2018graphvae}. MolGAN uses GANs to directly operate on graphs trained using RL objectives to generate output probabilities over an adjacency matrix and annotation matrix of a molecular graph \cite{de2018molgan}. However, these models suffer from scalability issues, which limit \textit{de novo} drug design to small molecule sizes to an atom length of ten \cite{li2018multi}.

\textbf{Fragment-based.} While atom-based generative models cover a wider chemical space, fragment-based approaches use a coarser molecular representation to constrain the search space \cite{meyers2021novo}. As a result, fragment-based models can maintain chemical validity during the generation process. The JT-VAE is a pioneering approach to fragment-based models \cite{jin2018junction}. First, the VAE generates a tree-structured scaffold of molecular substructures (fragment). Then, a graph message passing network decodes and assembles the fragments to construct the final molecular graph. In doing so, the JT-VAE can maintain molecular chemical validity in each step. You et al. \cite{gcpn} proposed GCPN that predicts a distribution of actions through a policy gradient algorithm to update the generated graph. 

\subsection{VAE models}
Deep generative models have recently shown great potential in \textit{de novo} drug design. RNNs and LSTM networks are often used to predict properties or generate sequences of characters, such as the SMILES notation. While GANs can produce new molecules from random noise input, they may not provide sufficient control over desired molecular properties. RL models define a reward function based on the desired molecular properties and find a policy that maximises the reward \cite{de2018molgan}. However, they may not provide a direct representation of generated structures. Therefore, GANs are often coupled with RL to generate molecules with controlled chemical properties. On the other hand, VAEs may provide a better balance between those deep generative models. VAEs for molecular generation can be considered either atom-based or fragment-based. This project aims to generate a unique set of novel molecules with specific properties using deep generative VAE models for \textit{de novo} drug design. 

\subsubsection{VAE}
VAEs have demonstrated remarkable success in various applications of generative modelling and unsupervised learning tasks. A VAE is an encoder-decoder neural network that learns to generate new data samples similar to a given training dataset by discovering a compact, low-dimensional data representation called the latent space \cite{doersch2016tutorial, pu2016variational}. The VAE observes a space $x$ through a prior distribution over a latent space $p(z)$ and a conditional likelihood of generating a data sample from a latent space $p_\theta(x|z)$. The encoder maps the input to the posterior density $q_\phi(z|x)$ over the latent variable $z$ with a multivariate Gaussian, $q_\phi(z|x)\sim\mathcal{N}(\mu_\phi,\sigma^2_\phi)$. The decoder then reconstructs the input data from the latent variable, given by the density $p_\theta(x|z)$. The VAE aims to learn the marginal log-likelihood of the observed data in the generative process. Since the log-likelihood is intractable, the evidence lower bound (ELBO) relies on the posterior distribution $q_\phi(z|x)$ to maximise the log-likelihood defined below:
\begin{equation}
\max_{\phi, \theta} \mathbb{E}_{q_\phi(z|x)}[log p_\theta(x|z)]\label{eq: vae pre1}
\end{equation}
where $\mathbb{E}$ is the expectation value \cite{richards2022conditional}. Equation (\ref{eq: vae pre1}) is also known as the ELBO formulation, where the VAE aims to minimise the reconstruction term for the encoder to generate meaningful latent vectors for the decoder to reconstruct. The optimisation objective of the VAE can be written as follows:
\begin{equation}
\mathcal{L}(\theta,\phi;x,z) = log p_\theta(x|z) - D_{KL}[q_\phi(z|x)\parallel p(z)]\label{eq: vae pre2}
\end{equation}
where $D_{KL}(\parallel)$ stands for non-negative Kullback-Leibler divergence loss between the true and approximate posterior \cite{higgins2017beta}. In short, the VAE aims to optimise $\theta$ and $\phi$ to minimise the reconstruction error between the input and the output and to make $q_\phi(z|x)$ as close as possible to $p_\theta(z|x)$, respectively. Since maximising the ELBO is equivalent to maximising the log-likelihood of the observed data and minimising the divergence of the approximate posterior from the exact posterior, Equations (\ref{eq: vae pre1}) and (\ref{eq: vae pre2}) can be rewritten as follows:
\begin{equation}
\begin{split}
log p_\theta(x|z) \geq & \mathcal{L}(\theta,\phi;x,z) \\= \mathbb{E}_{q_\phi(z|x)}[log p_\theta(x|z)]
& - D_{KL}[q_\phi(z|x)\parallel p(z)]. \label{eq: vae}
\end{split}
\end{equation}
However, VAEs may experience posterior collapse when the model learns a trivial local optimum of the ELBO objective where the variational posterior is misrepresented as the true posterior \cite{richards2022conditional}. Research by \cite{higgins2017beta, gomez2018automatic, burgess2018understanding} has shown that posterior collapse could be due to an insufficient representation of the reconstruction loss. During model training, VAEs assign equal weight to the reconstruction loss and $D_{KL}(\parallel)$ loss, causing the network to optimise $D_{KL}  (\parallel)$ prematurely. As a result, an inaccurate local optimum poorly approximates the true posterior.

\subsubsection{\texorpdfstring{$\beta$} -VAE} 
Higgins et al. \cite{higgins2017beta} proposed the $\beta$-VAE, which introduces a hyperparameter $\beta$ to the loss function to address posterior collapse. The $\beta$-VAE leverages the principles of VAE and disentangled representation learning to generate new molecules similar to the training dataset. Disentanglement refers to separating the latent variables in the VAE's latent space into independent dimensions corresponding to specific data aspects by multiplying the $D_{KL}(\parallel)$ loss in Equation (\ref{eq: vae}) by $\beta$ \cite{burgess2018understanding,mathieu2019disentangling}. 

While tuning $\beta$ is crucial in controlling the degree of disentanglement in the $\beta$-VAE, no general method currently quantifies the degree of learned disentanglement for \textit{de novo} drug design. Moreover, most literature aimed at tuning the value of $\beta$ automatically. Nonetheless, Higgins et al. suggested that a higher $\beta$ strengthens the constraint of the latent space to be disentangled, while a lower $\beta$ allows for greater flexibility in the representation. Equation (\ref{eq: bvae}) shows the resulting ELBO formulation for the objective function of the $\beta$-VAE:
\begin{equation}
\begin{split}
log p_\theta(x|z) \geq & \mathcal{L}(\theta,\phi;x,z) \\= \mathbb{E}_{q_\phi(z|x)}[log p_\theta(x|z)]
& - \beta D_{KL}[q_\phi(z|x)\parallel p(z)]. \label{eq: bvae}
\end{split}
\end{equation}

The $\beta$-VAE has been applied in \textit{de novo} drug discovery. For example, it can generate novel compounds with known biological activity by controlling the latent space dimensions associated with the desired activity using a training dataset of compounds with known activity \cite{gomez2018automatic}. Similarly, a $\beta$-VAE trained on a dataset of compounds with known physical properties, such as solubility or stability, can generate novel compounds with similar properties by controlling the associated latent space dimensions \cite{higgins2017beta, burgess2018understanding}. Recent research has shown that dynamically altering $\beta$ can lead to better results \cite{rydhmer2021dynamic}. Bowman et al. \cite{bowman2015generating} proposed a training scheduler that effectively disentangles latent representations in SMILES strings using a linear annealing process. However, it is essential to note that both VAE and $\beta$-VAE models may not generate molecules with desired properties.
\begin{figure*}
    \centering
    \includegraphics[width=0.8\textwidth]{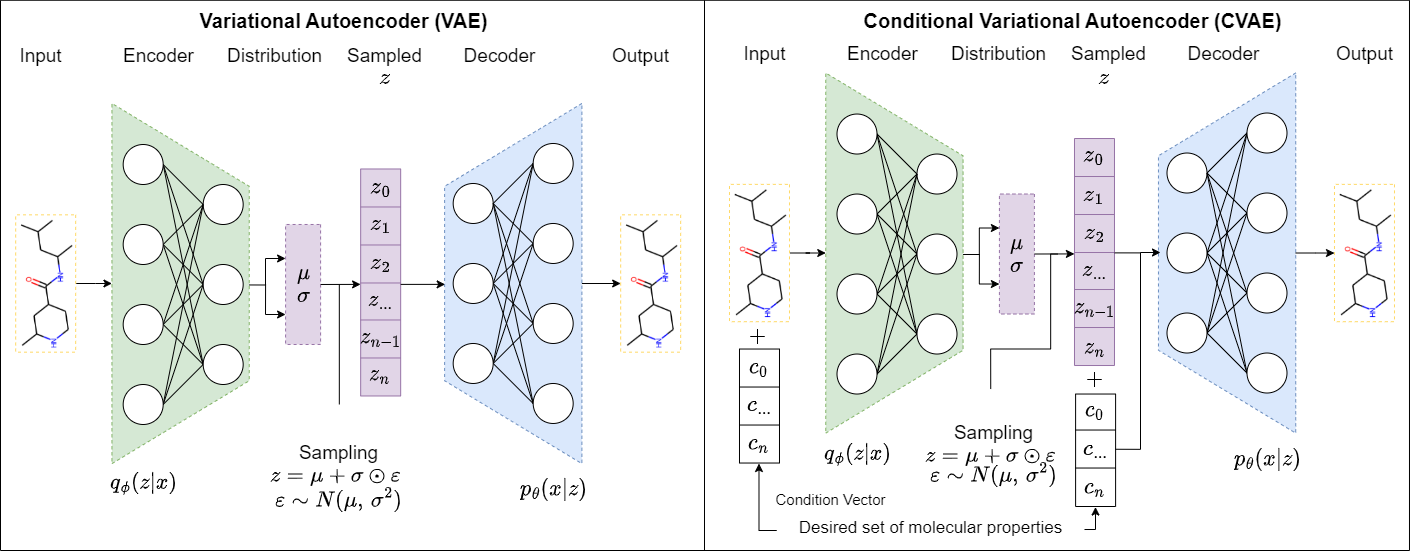}
    \caption{Left: A standard VAE reconstructs a new molecule similar to the input. Right: Latent conditioning is introduced as explicit inputs to a CVAE. The CVAE reconstructs a new molecule with desired and optimised molecular properties given the condition vector.}
    \label{fig:vae vs cvae}
\end{figure*}
\subsubsection{Conditional VAE} 
Explicit condition vectors can be introduced into the VAE's latent space to generate molecules with desired properties. This model is known as the conditional variational autoencoder (CVAE), which modifies the latent vectors based on the condition vector without adding network architecture or loss \cite{richards2022conditional, lim2018molecular}. As a result, the CVAE can control the latent space dimensions corresponding to target properties defined in the condition vector. Figure \ref{fig:vae vs cvae} depicts the model architecture difference between the VAE and the CVAE. Therefore, the ELBO objective function of the CVAE can be defined below: 
\begin{equation}
    \begin{split}
        \mathcal{L}(\theta,\phi;x,z, c) = \mathbb{E}_{q_\phi(z|x,c)}[log p_\theta(x|z),c] - D_{KL}[q_\phi(z|x,c)\parallel p(z|c)], \label{eq: cvae}
    \end{split}
\end{equation}
where the main difference with Equation (\ref{eq: bvae}) is the condition vector, $c$. The condition vector corresponds to the desired molecular properties that the model should learn when generating molecules. To that end, the decoder of the trained CVAE model could generate molecules with the specified properties using the condition and latent space vectors \cite{lee2022mgcvae}.

\subsection{Limitations} 
Although CVAEs have shown great success in generating molecules with desired properties through explicit latent conditioning, they may face challenges in effectively separating latent molecular representations, leading to posterior collapse similar to that of a VAE \cite{richards2022conditional}. To address this issue, we will introduce disentanglement in Equation (\ref{eq: cvae}) and empirically tune the $\beta$ hyperparameter.

\subsection{Contribution} 
The simultaneous optimisation of molecules with multiple properties remains a significant challenge in \textit{de novo} drug design. By introducing disentanglement to the CVAE, we aim to generate novel molecules with the following contribution: 
\begin{itemize}
    \item we introduced $\beta$ into the MGCVAE model \cite{lee2022mgcvae} to propose the $\beta$-CVAE model;
    \item we increased the maximum molecule length of 16 in contrast to previous research where small size molecules are ten or less;
    \item we applied multivariate explicit latent conditioning with ClogP, CMR, QED, and SAS as molecular properties;
    \item we empirically tuned the $\beta$ hyperparameter and analysed the quality of molecules generated;
    \item we applied a hyperoptimised gradient descent to train our models and discussed our findings \cite{gdtuo};
    \item we standardised evaluation metrics to evaluate the model performance based on the quality of molecules and multivariate molecular properties.
\end{itemize}

\section{Materials and methods}
\subsection{Dataset}
\begin{figure*}[h!]
    \centerline{\includegraphics[width=\textwidth]{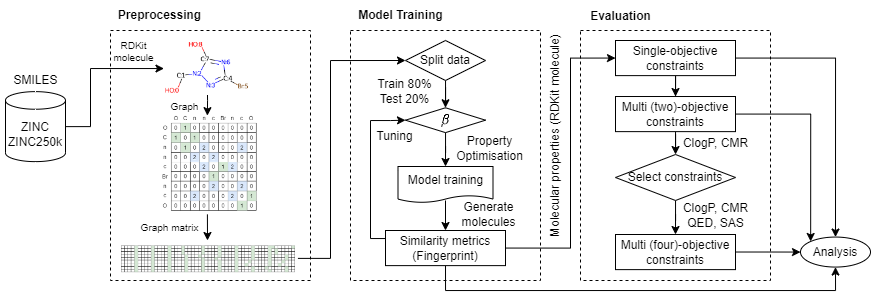}}
    \caption{Materials and methods pipeline.}
    \label{fig: methods}
\end{figure*}

We summarised our materials and methods in Figure \ref{fig: methods}. This study extracted a subset of 1.2 million molecules from 10 million molecules with 16 or fewer atoms (nodes) obtained from the ZINC database and ZINC-250k benchmark dataset. This was done to address scalability issues by including samples with longer sequences. In addition, we cleaned and removed SMILES with "$+$", "$-$", and  ".", which are not suitable for graph generation. We note that while ZINC molecules comprise 12 types of atoms (B, C, N, O, F, Si, P, S, Cl, Br, Sn, and I), the molecules in our dataset only comprise 8 types of atoms (C, N, O, F, Si, S, Cl, Br). This may generate molecules without B, P, Sn, and I elements since the VAE models aim to generate molecules that resemble the dataset. The sampled molecules have four types of bonds (single, double, triple, and aromatic).

\clearpage
\subsection{Preprocessing}
To generate molecular graphs, we used RDKit to convert SMILES into molecules \cite{landrum2013rdkit}. Subsequently, we adopted the initial graph representation based on the encoding process from Lee and Min \cite{lee2022mgcvae}. This representation method allows each initial graph matrix to be constructed uniquely \cite{lee2022mgcvae}. Furthermore, the initial graph matrix is a more sophisticated form of graph-based representation that attempts to capture the complexity of the molecular structure. The generated molecules can also be easily reconstructed and converted into SMILES for post-processing and evaluation. Finally, the transformed dataset was divided into training and test sets at a ratio of 8:2. 
\begin{figure}[t!]
    \centering
    \includegraphics[width=0.9\textwidth]{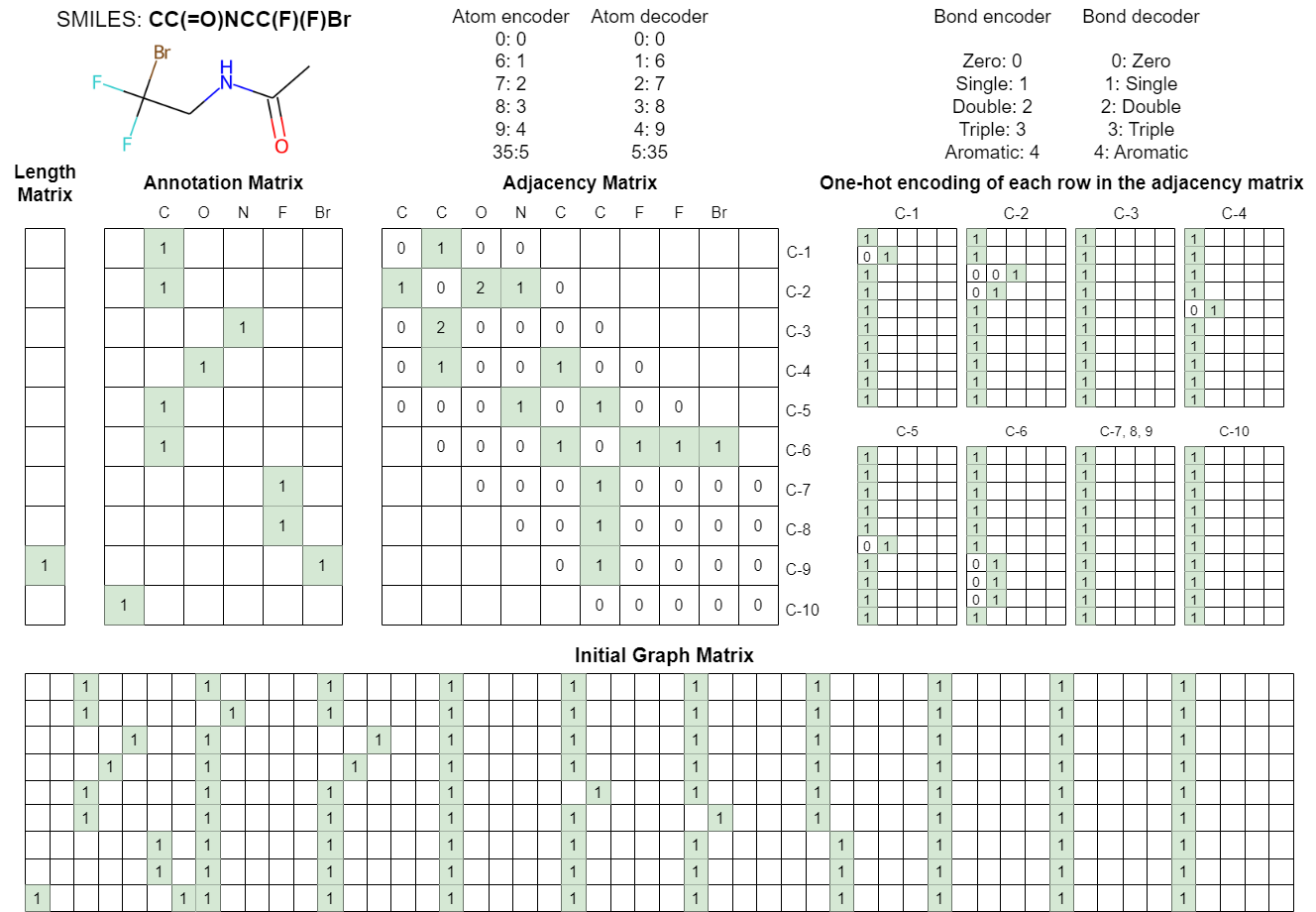}
    \caption{Construction of the initial graph matrix for CC(=O)NCC(F)(F)Br. The length matrix encodes the length of the molecule. Here, we use 10 to demonstrate padding in the construction. The annotation matrix encodes each atom, and the adjacency matrix describes the bond information \cite{lee2022mgcvae}. To reduce the computational intensity, we consider the upper right triangle of the adjacency matrix (symmetry) to construct the one-hot encoding of each row in the adjacency matrix.}
    \label{fig:construction}
\end{figure}

\subsection{Model}
\begin{figure}[t!]
    \centering
    \includegraphics[width=0.6\textwidth]{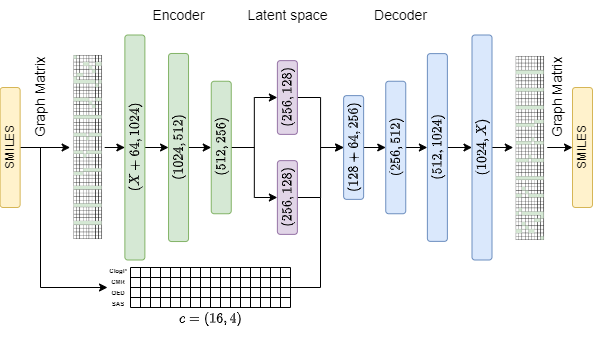}
    \caption{$\beta$-CVAE model architecture.}
    \label{fig: model architecture}
\end{figure}

We proposed the $\beta$-conditional variational autoencoder ($\beta$-CVAE) in this study. The $\beta$-CVAE is an extension of the aforementioned VAEs, specifically the MGCVAE \cite{lee2022mgcvae}. We rewrite Equation (\ref{eq: cvae}) to define the objective function of $\beta$-CVAE as follows:
\begin{equation}
    \begin{split}
        \mathcal{L}(\theta,\phi;x,z, c) = \mathbb{E}_{q_\phi(z|x,c)}[log p_\theta(x|z),c] -  \beta D_{KL}[q_\phi(z|x,c)\parallel p(z|c)],\label{eq: bcvae ch3}
    \end{split}
\end{equation}

where $\beta$ is the hyperparameter to introduce disentanglement. Figure \ref{fig: model architecture} shows our proposed model architecture. We employed a conventional bottleneck VAE framework with two hidden layers of 1024 and 512 dimensions and set the latent space dimension as 128. In our study, we excluded dropout and batch normalisation due to the following reasons: 1) dropout masks the inputs during training, which distorts the latent space and affects the reconstruction loss; 2) batch normalisation modifies the mean and variance of the prior distribution, which may lead to poor performance. In Equation (\ref{eq: bcvae ch3}), $\beta$$D_{KL}(\parallel)$ can be considered a form of regularisation. We explored the qualitative effects of disentanglement by tuning the value of $\beta$ and evaluated the properties of the generated molecules. We began with $\beta = 1$, where the model is reduced to a regular CVAE. Subsequently, we empirically tuned the value of $\beta \in [0.01, 10]$ and compared the quality of molecules generated based on the desired properties. 

\subsection{Multivariate property optimisation}
To generate molecules based on the Ghose filter, we introduced ClogP and CMR into the latent space of the $\beta$-CVAE as a pairwise condition vector. The Ghose filter suggests that drug-like molecules have ClogP $\in [-0.4, 5.6]$ or CMR $\in [40, 130]$ \cite{ghose1999knowledge}. We discretised the Ghose filter into C1 and C2, and defined the desired molecular properties in Equation (\ref{eq: multi summary}):
\begin{equation}
    \begin{aligned}
    & C1 = ClogP = \{0, 1, 2, 3, 4, 5\} \\
    & C2 = CMR  = 0.1\{40, 50, 60, 70, 80\} \\
    & C3 = QED = 10\{0.5, 0.6, 0.7, 0.8, 0.9\} \\
    & C4 = SAS  = \{3, 4, 5, 6\}, \label{eq: multi summary}
    \end{aligned}
\end{equation}
where C3 and C4 are the other molecular properties we want to optimise. CMR and QED are scaled by 0.1 and 10, respectively \cite{lee2022mgcvae}. We chose a high QED and SAS condition vector to experiment whether the model can improve the quality of molecules generated (exploitation). 

\subsection{Network training}
We implemented all models using PyTorch and conducted training locally on an NVIDIA GeForce RTX2060 GPU with 16GB of memory. The models were trained by reading the training dataset with a batch size of 256 initial graph matrices in one training epoch. Table \ref{tab: network training} provides a summary of training hyperparameters, where we consider $\beta$ as part of a training hyperparameter since it modifies the loss function defined in Equation (\ref{eq: bcvae ch3}). To address the tedious task of tuning the optimiser's hyperparameters manually, we utilised a modified gradient-based optimisation algorithm known as \textit{gradient descent: the ultimate optimiser} (\textit{gdtuo}) proposed by Chandra et al. \cite{gdtuo}. The authors modified the backpropagation algorithm to compute "hyper-gradients" to optimise the hyperparameters alongside the model parameters during training rather than manually choosing beforehand. Using the \textit{gdtuo} optimiser, we expect the quality of models generated to improve since the training process is automated and optimised. We used the same hyperparameters for both optimiser training described in Table \ref{tab: network training}. 

\begin{table}[h!]
\caption{Network training.}
\centering
        \begin{tabular}{ll}
        \toprule[1pt]
        Hyperparameter & Value \\ \midrule[0.5pt]
        Optimiser type & Adam and Adam \textit{gdtuo} \\
        Learning rate & 0.005 (Adam) \\
        Number of epochs & 100 (Adam and \textit{gdtuo}) \\
        Batch size & 256 (Adam and \textit{gdtuo}) \\
        $\beta$ & 0.01; 0.1; 0.5; 1.0; 2.0; 5.0, 10.0 \\ \bottomrule[1pt]
        \end{tabular}
        \label{tab: network training}
\end{table}

\subsection{Evaluation methods}
We evaluated our model based on De Cao and Kipf \cite{de2018molgan} and the quality of molecules generated through similarity metrics defined in Samanta et al. \cite{samanta2020nevae}. All performance metrics were reported to two decimal places. We then drew conclusions about the representation method used in this study, which was not evaluated in \cite{lee2022mgcvae}.

\subsubsection{Model performance}
\begin{figure}[h!]
    \centering
    \includegraphics[width=0.7\textwidth]{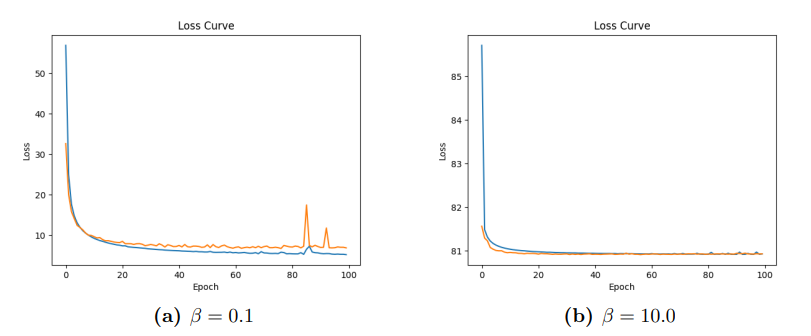}
    \caption{Training loss curves.}%
    \label{fig:loss}
\end{figure}

We tested the $\beta$-CVAE models on reconstructing input graph matrices from their latent representations and decoding them when sampling from the prior distribution. To empirically determine $\beta$, we compared the quality of molecules generated based on similarity metrics. This step is crucial for our subsequent analyses to compare with the baseline models: VAE, $\beta$-VAE, and CVAE using the same representation method. With reference to Figure \ref{fig:loss}, it is important to note that we do not evaluate the magnitude of the loss functions. Naturally, a smaller $\beta = 0.01$ gives rise to a smaller training loss than $\beta = 0.1$ as it multiplies with $D_{KL}(\parallel)$. Hence, we only used the training loss curve to ensure the model training converges. Therefore, evaluating our model performance by analysing the quality of generated molecules through similarity metrics and molecule property optimisation was imperative. 

\subsubsection{Similarity metrics}
We assessed the performance of our model based on the similarity metrics: validity, novelty, and uniqueness metrics \cite{richards2022conditional,lee2022mgcvae,2017rnn, ORGAN}. Validity is defined as V = $100\times|C_s|/n_s$, where $C_s$ is the set of validly generated molecules and $n_s$ is the total number of generated molecules. Secondly, novelty is defined as N(0.9) = $100\times(1-|C_s\cap D|/|C_s|)$, where $D$ is the training dataset. Lastly, uniqueness is defined U(0.9)=$100\times set(C_S)/n_s$. Using these similarity metrics, we compared our model with baseline models using different values of $\beta$. We then chose the value of $\beta$ that generated the highest quality of molecules for molecular property evaluation.

\subsubsection{Evaluation of \texorpdfstring{$\beta$}{}} 
To evaluate the effects of disentanglement, we compared our model with baseline models and with various values of $\beta$. Using the similarity metrics, we chose the value of $\beta$ that generated the highest quality of molecules for further evaluation. Subsequently, we evaluated the uniqueness metric based on the different $\beta$ values and reported our findings.

\subsubsection{Molecular property optimisation}
We divided our analysis into single-objective (univariate) property optimisation and multi-objective (multivariate) property optimisation. For the primary analysis, we first considered C1 and C2 in Equation (\ref{eq: multi summary}) and experimented with different $\beta$ values. Using similarity metrics, we studied the effects of disentanglement with different $\beta$. We classified that a molecular property is within the range of the conditional value if $C1 \pm 0.5$ and $C2 \pm 5$. Similarly, we simultaneously evaluated the number of molecules that satisfied the pairwise conditions. The top three pairwise conditions were subjected to further classification defined by $C3 \pm 0.05$ and $C4 \pm 0.5$. We analysed how introducing QED and SAS modify the quality of molecules generated.

\subsection{Optimiser comparison}
Finally, we compared PyTorch's Adam optimiser and the hyperoptimised \textit{gdtuo} optimiser \cite{gdtuo}. Since there is no metric evaluating the performance between the two in the context of drug discovery, we analysed the quality of molecules (similarity metrics) and the molecular properties (molecular property optimisation) of the molecules generated. 

\section{Results}
\subsection{Model performance} 
We set the non-conditional models (VAE and $\beta$-VAE) to generate 1000 molecules and 1000 molecules per condition for the conditional models (CVAE and $\beta$-CVAE). Here, we present the experimental results from tuning the $\beta$ hyperparameter. We first report the quantitative analysis of our model (shaded) in Table \ref{tab: quality of molecules for beta} by comparing the quality of the molecules generated with the baseline models. Then, we present the scores for the conditional models (CVAE and $\beta$-CVAE) by calculating the average quality across all generated molecules. Our model generated 100\% valid molecules. Despite the strict threshold values for novelty and uniqueness, our model produced 100\% novel molecules and competitive uniqueness scores to \cite{richards2022conditional,lee2022mgcvae, molecularrnn}. 

\begin{table}[h!]
\centering
\caption{Quality of generated molecules.}
\begin{tabular}{llll}
\toprule[1pt]
Algorithm & Validity & Novelty & Uniqueness \\ \midrule[0.5pt]
VAE & 99.27$\pm$0.01  & 100.00 & 93.17$\pm$0.01  \\
$\beta$-VAE ($\beta$=0.01) & 98.52$\pm$0.01 & 100.00 & 91.52$\pm$0.01 \\
$\beta$-VAE ($\beta$=0.1) & 98.58$\pm$0.01 & 100.00 & 89.62$\pm$0.01 \\
$\beta$-VAE ($\beta$=0.5) & 98.89$\pm$0.01 & 100.00 & 85.80$\pm$0.01 \\	
CVAE ($\beta$-CVAE 1.0) & 100.00 & 100.00 & 91.25$\pm$0.01 \\ \midrule[0.25pt]
\cellcolor[HTML]{EFEFEF}$\beta$-CVAE ($\beta$=0.01) & \textbf{100.00} & \textbf{100.00} & \textbf{99.44$\pm$0.01} \\
\cellcolor[HTML]{EFEFEF}$\beta$-CVAE ($\beta$=0.1) & 100.00 & 100.00 & 95.31$\pm$0.01 \\
\cellcolor[HTML]{EFEFEF}$\beta$-CVAE ($\beta$=0.5) & 100.00 & 100.00 & 93.07$\pm$0.01 \\
\cellcolor[HTML]{EFEFEF}$\beta$-CVAE ($\beta$=2.0) & 100.00 & 100.00 & 77.98$\pm$0.01 \\
\cellcolor[HTML]{EFEFEF}$\beta$-CVAE ($\beta$=10.0) & 100.00 & 100.00 & 46.54$\pm$0.01 \\ \bottomrule[1pt]
\end{tabular}
\label{tab: quality of molecules for beta}
\end{table}

\subsection{Evaluating \texorpdfstring{$\beta$}{}}
\begin{figure}[h!]
    \centering
    \includegraphics[width=0.5\textwidth]{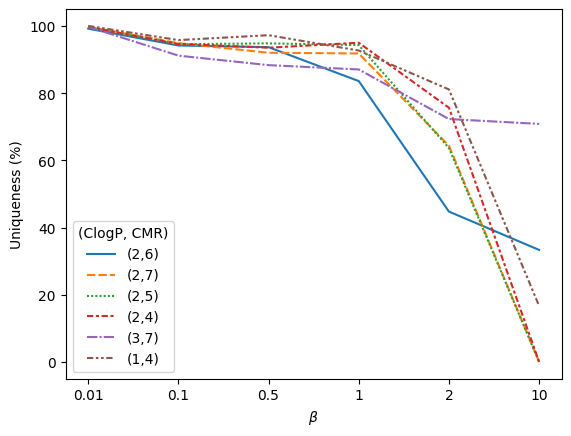}
    \caption{Uniqueness (\%) against $\beta$ values.}
    \label{fig:quality against beta}
\end{figure}

From Table \ref{tab: quality of molecules for beta} and Figure \ref{fig:quality against beta}, the quality of uniqueness decreases as the value of disentanglement ($\beta$) increases. Both $\beta$-VAE and $\beta$-CVAE displayed the same trend. This suggests that a lower $\beta$ value generates an increasingly diverse set of molecules. Furthermore, our model displayed improvement across all similarity metrics compared to the string-based $\beta$-CVAE presented in \cite{richards2022conditional}. Finally, this suggests that the graph matrix introduced by \cite{lee2022mgcvae} uniquely represents molecules, thereby improving the quality of generated molecules.

\textbf{Analysing $\beta$}\textbf{.}
In disentangled generation, a higher disentanglement factor penalises the model for learning highly correlated features with the dataset. However, we discovered that even with a low $\beta$ of 0.01, we could still generate 100\% novel molecules. In image generation tasks, noise is introduced in the training process to generate a similar but novel image. However, there is a limit beyond which the image stops being visually plausible and unrecognisable. Hence, the Kullback–Leibler divergence ($D_{KL}(\parallel)$) comes in to reconstruct the image plausibly. 

We selected $\beta = 0.01$ and expected our model to ignore the latent distribution. As a result, our model generated 100\% novel molecules. However, we should also expect poorer uniqueness scores, as a smaller value of $\beta$ should favour a correlated dataset distribution, including molecular properties. In contrast, our model generated more unique molecules without compromising molecular property optimisation tasks. This suggests that introducing the explicit latent conditioning vector to the graph matrix representation was not compromised with a lower value of $\beta$. Instead, it was able to generate [balance] a unique set of novel molecules [exploration] with desired properties [exploitation].

\subsection{Molecular property optimisation}
To further evaluate $\beta$, we chose the top three pairwise condition vectors to compare the quality of molecules generated from C1 and C2. Since our model generated 100\% valid and novel molecules, we only reported the quality of uniqueness in Table \ref{tab: quality of unique mols given cond vec} and plotted the trend in Figure \ref{fig:quality against beta}. We can observe that the uniqueness between generated molecules drops as we increase the disentanglement value. This result is consistent with Table \ref{tab: quality of molecules for beta}.

\begin{table}[h!]
\centering
\caption{Quality of unique molecules given top three condition vectors.}
\begin{tabular}{cccccccc}
\toprule[1pt]
\multicolumn{2}{c}{\textbf{Conditions}} & \multicolumn{6}{c}{\textbf{Uniqueness ($\% \pm$ 0.01\%})}\\ \midrule[0.5pt]
CLogP & CMR & \cellcolor[HTML]{EFEFEF}$\beta$=0.01& $\beta$=0.1 & $\beta$=0.5 & $\beta$=1 & $\beta$=2 & $\beta$=10  \\
2 & 60 & \cellcolor[HTML]{EFEFEF}\textbf{99.23} & 94.18 & 93.68 & 83.59 & 44.75 & 33.33 \\
2 & 70 & \cellcolor[HTML]{EFEFEF}\textbf{99.77} & 94.881 & 92.01 & 91.81 & 64.35 & 0 \\
3 & 70 & \cellcolor[HTML]{EFEFEF}\textbf{99.74} & 91.16 & 88.33 & 87.04 & 72.29 & 70.83 \\ \bottomrule[1pt]
\end{tabular}
\label{tab: quality of unique mols given cond vec}
\end{table}

\subsection{Single-objective optimisation}
Using $\beta$(=0.01)-CVAE, we present our results in Table \ref{tab: mol properties within ghose}. For ClogP, our model generated \textbf{99.73\% $\pm$ 0.01\%} (\textbf{4.93\% $\pm$ 0.01\% increase} from dataset) of molecules within the Ghose filter. For CMR, our model generated \textbf{99.98\% $\pm$ 0.01\%} of molecules within the Ghose filter (\textbf{1.98\% $\pm$ 0.01\% increase} from dataset). Compared to the non-conditioned baseline models, the conditioned models generated more molecules that satisfy the Ghose filter, thus suggesting that explicit latent conditioning generates molecules with desired ClogP or CMR values. Our findings are consistent with previous studies in \cite{richards2022conditional, meyers2021novo, li2018multi, lim2018molecular}. 

Interestingly, we found that the models with disentanglement generated a higher ratio of molecules out of the Ghose filter than the non-disentangled models. This is likely due to the lower disentanglement value of $\beta = 0.01$ used for our evaluation and analysis, which prompted exploration (instead of exploitation) of the latent space. In terms of single-objective optimisation scores, our model generated an average of \textbf{42.15\% $\pm$ 0.01\%} and \textbf{61.91\% $\pm$ 0.01\%} of molecules that satisfied C1 (ClogP) = 2 and C2 (CMR) = 60, respectively. Furthermore, our model scored the best among the baseline models by comparing the single-property scores in Table \ref{tab: comparison btwn models for c1 and c2}. Therefore, our results suggest that explicit latent conditioning plays an essential role in exploitation for \textit{de novo} drug design. Our model optimised molecular properties (exploitation) while improving the uniqueness of molecules generated (exploration). In other words, our model suggests balancing exploration and exploration to achieve desired compounds. 

\begin{table}[h!]
\caption{Molecular properties within Ghose filter.}
\centering
\begin{tabular}{cccccccc}
    \toprule[1pt]
    \multicolumn{4}{c}{\textbf{ClogP ($\% \pm$ 0.01\%)}} & \multicolumn{4}{c}{\textbf{CMR ($\% \pm$ 0.01\%)}} \\ \midrule[0.5pt]
    VAE & $\beta$-VAE & CVAE & \cellcolor[HTML]{EFEFEF}\textbf{$\beta$-CVAE} & VAE & $\beta$-VAE & CVAE & \cellcolor[HTML]{EFEFEF}\textbf{$\beta$-CVAE} \\
    98.78 & 98.29 & 99.94 & \cellcolor[HTML]{EFEFEF}99.73 & 98.54 & 98.01 & 100.00 & \cellcolor[HTML]{EFEFEF} 99.98 \\ \bottomrule[1pt]
    \end{tabular}
    \label{tab: mol properties within ghose}
\end{table}

\subsection{Multi-objective optimisation}
Continuing from our analysis and referring to Table \ref{tab: comparison btwn models for c1 and c2}, we also found our model generated 30.07\% $\pm$ 0.01\% of molecules that satisfy both properties simultaneously. We now assess four molecular properties simultaneously. 

\begin{table}[h!]
\caption{Comparison between models for C1 = 2 and C2 = 60 single-objective optimisation and multi-objective optimisation.}
\centering
\begin{tabular}{cccccc}
\toprule[1pt]
\multicolumn{2}{c}{\textbf{Optimisation}} & \multicolumn{4}{c}{\textbf{Model scores ($\% \pm$ 0.01\%)}} \\ \midrule[0.75pt]
Property & Condition & VAE & $\beta$-VAE & CVAE & \cellcolor[HTML]{EFEFEF}$\beta$-CVAE \\ \midrule[0.5pt]
ClogP & 2 & 36.10 & 32.71 & 41.07 & \cellcolor[HTML]{EFEFEF}\textbf{42.15} \\
CMR & 60 & 59.76 & 49.39 & 61.91 & \cellcolor[HTML]{EFEFEF}\textbf{66.76} \\ \midrule[0.25pt]
\multicolumn{2}{c}{Both conditions \{2, 60\}} & {25.98} & {l}{20.05} & {29.78} & \cellcolor[HTML]{EFEFEF}{\textbf{30.07}} \\ \bottomrule[1pt]
\end{tabular} 
\label{tab: comparison btwn models for c1 and c2}
\end{table}

\begin{figure}[h!]
    \centering
    \includegraphics[width=0.7\textwidth]{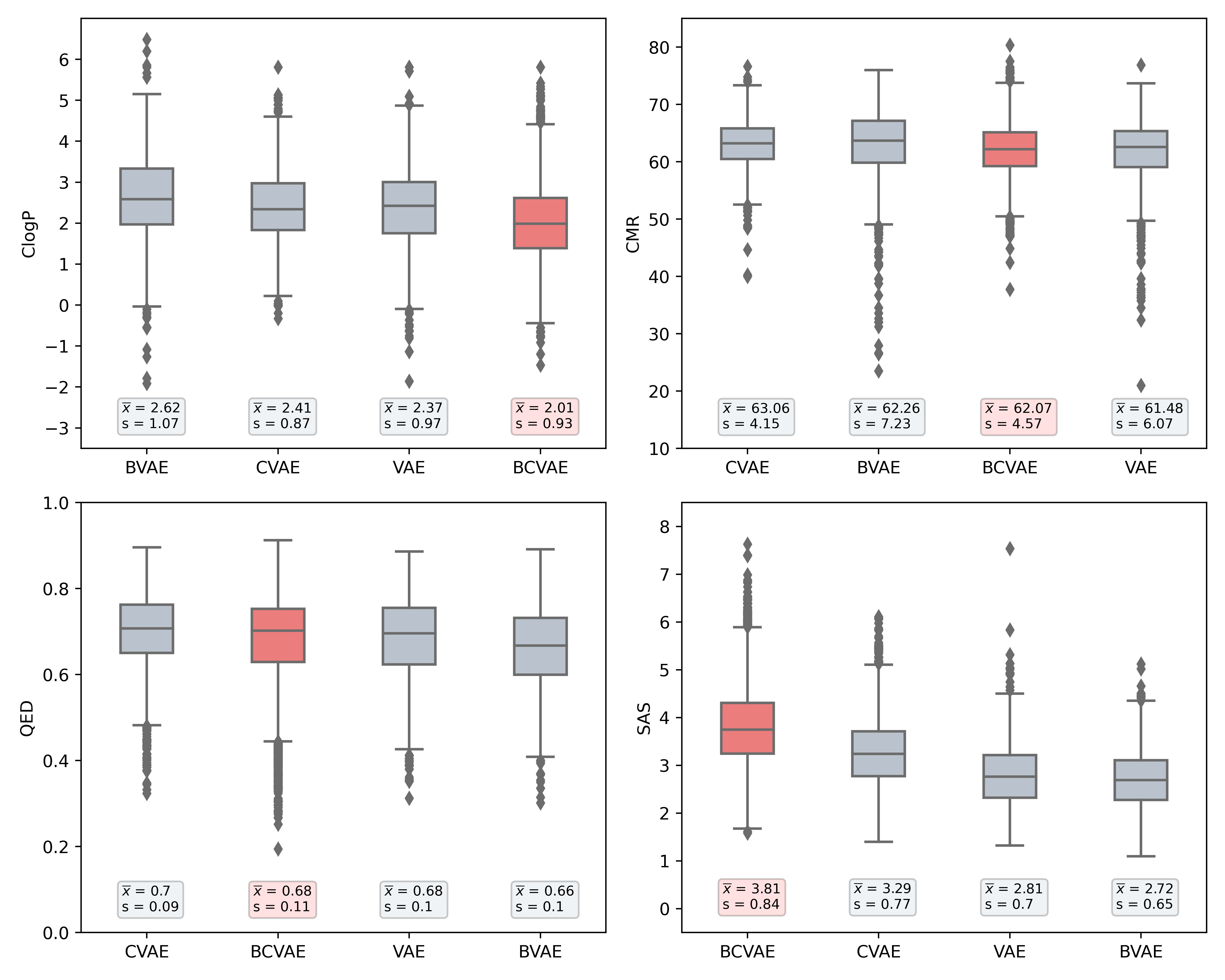}
    \caption{Molecular property subplots against models ordered in decreasing mean.}
    \label{fig:descriptive sum}
\end{figure}

We summarise the distribution of molecular properties in Figure \ref{fig:descriptive sum}, where we ordered the boxplots in decreasing mean with our model highlighted in red. We aim to exploit the latent space for optimised molecular properties for QED and SAS, where a higher mean score for both QED and SAS reflects better model performance. For QED, our model ranked second behind the CVAE due to the presence of outliers. A possible explanation for the lower mean could be related to the high uniqueness among the molecules generated, which we deemed acceptable previously. For SAS, our model generated the highest mean value, which reflects that our generated molecules are easier to synthesise than the baseline models. It improved the ease of synthesis by at least \textbf{40.07\% $\pm$ 0.01\%} compared to the VAE. In this case, we favour the outliers with high SAS.

We present the top 2 combination scores where all four desired properties are satisfied. For the optimisation of \{C1, C2, C3, C4\} = \{2, 60, 0.7, 4\}, our model generated \{43.68\% $\pm$ 0.01\%, 68.02\% $\pm$ 0.01\%, \textbf{98.93\% $\pm$ 0.01\%}, \textbf{43.20\%$\pm$ 0.01\%}\} of molecules that satisfied all desired properties. For the optimisation of \{C1, C2, C3, C4\} = \{3, 60, 0.9, 4\}, our model generated \{23.15\% $\pm$ 0.01\%, 68.02\% $\pm$ 0.01\%, \textbf{100.00\% $\pm$ 0.01\%}, \textbf{43.20\% $\pm$ 0.01\%}\} of molecules that satisfied all desired properties. We emphasise that generating molecules satisfying QED and SAS is an exploitation task where physiochemical trade-offs between other molecular properties could exist. Nonetheless, these results suggest that our model can optimise multi-objective (four) properties simultaneously. 

\begin{figure}[h!]
    \centering
    \includegraphics[width=0.8\textwidth]{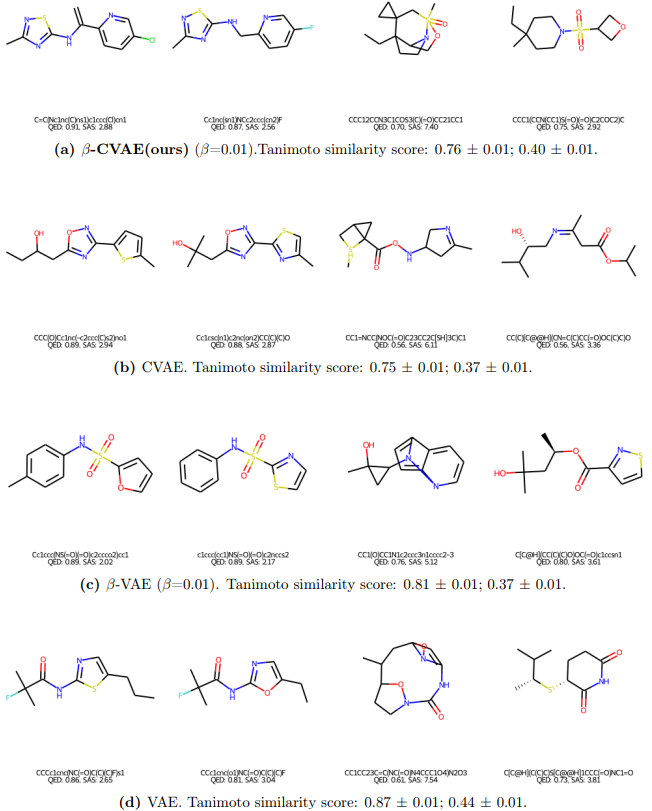}
    \caption{Molecular similarity analysis: our model (a) against the baseline models (b)-(d). From left to right: highest QED generated, most similar (dataset), highest SAS generated, most similar (dataset) \cite{landrum2013rdkit}.}%
    \label{fig: mol similarity analysis}%
\end{figure}

\begin{figure}[h!]
    \centering
    \includegraphics[width=0.8\textwidth]{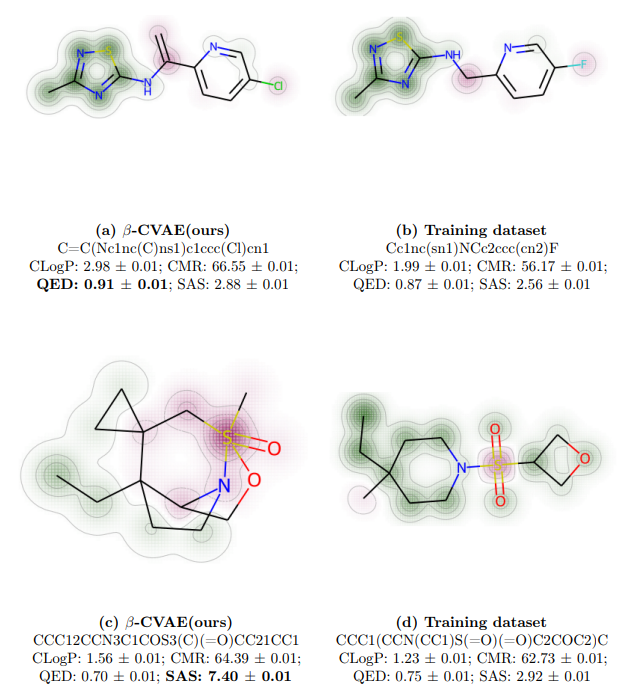}
    \caption{Closer look at Figure \ref{fig: mol similarity analysis}(a) through similarity mapping. Colour scheme: gradient of similarity (green), no change in similarity (grey), gradient of dissimilarity (pink) \cite{mapping}.}%
    \label{fig:closer look}
\end{figure}

To visualise the generated molecules with the most similar molecule in the training dataset, we calculated the Tanimoto similarity using ECFP4 fingerprints. The leftmost molecule (ours) in Figure \ref{fig: mol similarity analysis}(a) optimised on QED achieved \textbf{QED = 0.91 $\pm$ 0.01}. The molecule to the right is the most similar dataset molecule, with QED = 0.87 $\pm$ 0.01 and a similarity score of 0.76 $\pm$ 0.01. The third molecule (ours) was optimised on SAS and achieved a score of \textbf{7.63 $\pm$ 0.01}. In contrast, the most similar dataset molecule (rightmost molecule) has SAS = 3.92 $\pm$ 0.01, with a similarity score of 0.40 $\pm$ 0.01. Finally, we present a closer look at the molecules in Figure \ref{fig: mol similarity analysis}(a) with similarity mapping in Figure \ref{fig:closer look}. Both figures suggest our model can generate unique molecules with multivariate optimised properties. 

In summary, our model generated highly unique molecules in terms of substructures while generating molecules with desired properties through multivariate property optimisation. Our model could balance the trade-offs resulting from a low Kullback-Leibler divergence loss, where $\beta \in [0.01, 0.1]$ provides a good balance between exploration (generating unique molecules) and exploitation (optimising molecular properties).

\subsection{Optimiser comparison}
Using $\beta$-CVAE = 0.01, we assess the quality of molecules and the molecular property optimisation for PyTorch's Adam optimiser and Chandra et al.'s \textit{gdtuo} \cite{gdtuo}. While both model training converges, we report that the \textit{gdtuo} optimiser took an average of 4.76\% $\pm$ 0.01\% in additional runtime than PyTorch's Adam optimiser. Table \ref{tab: optimiser comparison} summarises the results. We found that the hyperoptimised Adam \textit{gdtuo} optimiser outperforms the regular Adam optimiser in our evaluation methods - the quality of molecules (uniqueness) and molecular property optimisation. We note that while the authors in \cite{gdtuo} evaluated their results on benchmarked convolutional neural networks (CNNs) and RNNs, our results concur with them that hyperoptimised Adam was beneficial for a more extensive regime and complex real-world task of \textit{de novo} drug design like ours. 

\begin{table}[h!]
\caption{Optimiser comparison between PyTorch's Adam and Hyperoptimised Adam.}
\centering
\begin{tabular}{cccccc}
\toprule[1pt]
\multicolumn{2}{c}{\textbf{Conditions}} & \multicolumn{4}{c}{\textbf{$\beta$-CVAE(Adam)} $\approx$ 168 minutes} \\ \midrule[0.5pt]
ClogP & CMR & \multicolumn{2}{c}{Single-objective} & Multi-objective & Uniqueness \\ \midrule[0.25pt]
1 & 60 & 29.54 & 61.78 & 22.01 & 99.75 \\
2 & 60 & 37.71 & 61.41 & 23.92 & \textbf{99.71} \\
3 & 70 & 34.54 & 62.54 & 23.84 & 99.17 \\ \midrule[0.75pt]
\multicolumn{2}{c}{\textbf{Conditions}} & \multicolumn{4}{c}{\textbf{$\beta$-CVAE(gdtuo)} $\approx$ 176 minutes} \\ \midrule[0.5pt]
ClogP & CMR & \multicolumn{2}{c}{Single-objective} & Multi-objective & Uniqueness \\ \midrule[0.25pt]
1 & 60 & 32.15 & 70.71 & \textbf{25.80} & \textbf{99.82} \\
2 & 60 & 42.15 & 66.76 & \textbf{30.07} & 99.23 \\
3 & 70 & 34.85 & 64.61 & \textbf{25.71} & \textbf{99.77} \\ \bottomrule[1pt]
\end{tabular} \label{tab: optimiser comparison}
\end{table}

\section{Discussion}
Results suggest that our model is suitable for single-objective and multi-objective optimisation of generating molecules with desired properties. Furthermore, our model allows us to balance exploration and exploitation through disentanglement and explicit latent space conditioning.

\subsection{Limitations}
\textbf{Dataset.} In this study, we extracted molecules from the ZINC database, which is a database of commercially available compounds \cite{irwin2005zinc}. After filtering for molecules with less than 17 atoms (nodes), we found that our dataset contains only eight types of atoms. As a result, our model generated molecules that only contain these eight types of atoms due to the dataset constraints. This possibly limited our models to generate less sophisticated compounds using more types of atoms (B, P, Sn, and I). 

\subsection{Benchmarking our method}
We aim to evaluate our results with Table \ref{tab: summary of related work}. In addition, we aim to benchmark our method using the QM9 and ChEMBL databases \cite{chembl,qm9}. The QM9 database provides quantum chemical properties for small organic molecules, and the ChEMBL database is a manually curated database of bioactive molecules with drug-like properties. We could also benchmark our model against previous work if the model is available in online repositories. 

\subsection{\texorpdfstring{$\beta$}{}-models}
In this work, we empirically evaluated our model with different values of $\beta$. We also could explore dynamic $\beta$-VAE suggested in \cite{rydhmer2021dynamic} subjected to conditional information. Similarly, \cite{shao2020controlvae, shao2021controlvae} suggested tuning the VAE using control theory. A possible direction could involve the design of self-tuning control schemes for $\beta$ tuning using neural networks instead of Proportional-Integral controllers for VAE in the context of \textit{de novo} molecular generation \cite{ang2018proposed}. However, we highlight that our main contribution is analysing the quality of molecules generated using different $\beta$ values, which is absent in literature regarding automating the $\beta$ tuning processes. Having analysed the effects of $\beta$, we are in a better position to explore dynamically tuning $\beta$ \cite{rydhmer2021dynamic,shao2020controlvae, shao2021controlvae, ang2018proposed}. 

\section{Conclusion}
In conclusion, we proposed a $\beta$-CVAE deep generative model for \textit{de novo} drug design. We introduced multivariate explicit latent conditioning to the initial graph matrix and employed a hyperoptimised optimiser for network training and standardised evaluation methods. We empirically studied the effects of disentanglement on the quality of molecules and molecular property optimisation. Results showed that our model generated unique molecules with optimised drug-related properties when $\beta \in [0.01, 0.1]$, suggesting that our model could balance exploration and exploitation. Furthermore, our model outperformed baseline models and remained competitive with previous work. Future studies could include larger molecule sizes comprising H, B, P, Sn, and I elements. Therefore, we have demonstrated the potential of our model in the application of multivariate optimisation for \textit{de novo} drug design. We are thus in an excellent position to begin further research. 

\section*{Acknowledgements}
We thank Associate Professor Erik Birgersson for his guidance and for providing valuable advice. 

\clearpage
\bibliographystyle{unsrt}  
\bibliography{references}  

\begin{thebibliography}{10}

\bibitem{polishchuk2013estimation}
Pavel~G Polishchuk, Timur~I Madzhidov, and Alexandre Varnek.
\newblock Estimation of the size of drug-like chemical space based on gdb-17
  data.
\newblock {\em Journal of computer-aided molecular design}, 27(8):675--679,
  2013.

\bibitem{kirkpatrick2004chemical}
Peter Kirkpatrick and Clare Ellis.
\newblock Chemical space.
\newblock {\em Nature}, 432(7019):823--824, 2004.

\bibitem{devi2015evolutionary}
R~Vasundhara Devi, S~Siva Sathya, and Mohane~Selvaraj Coumar.
\newblock Evolutionary algorithms for de novo drug design--a survey.
\newblock {\em Applied Soft Computing}, 27:543--552, 2015.

\bibitem{richards2022conditional}
Ryan~J Richards and Austen~M Groener.
\newblock Conditional $\beta$-vae for de novo molecular generation.
\newblock {\em arXiv preprint arXiv:2205.01592}, 2022.

\bibitem{higgins2017beta}
Irina Higgins, Loic Matthey, Arka Pal, Christopher Burgess, Xavier Glorot,
  Matthew Botvinick, Shakir Mohamed, and Alexander Lerchner.
\newblock beta-vae: Learning basic visual concepts with a constrained
  variational framework.
\newblock In {\em International conference on learning representations}, 2017.

\bibitem{lee2022mgcvae}
Myeonghun Lee and Kyoungmin Min.
\newblock Mgcvae: Multi-objective inverse design via molecular graph
  conditional variational autoencoder.
\newblock {\em Journal of Chemical Information and Modeling}, 2022.

\bibitem{weininger1988smiles}
David Weininger.
\newblock Smiles, a chemical language and information system. 1. introduction
  to methodology and encoding rules.
\newblock {\em Journal of chemical information and computer sciences},
  28(1):31--36, 1988.

\bibitem{carracedo2021review}
Paula Carracedo-Reboredo, Jose Li{\~n}ares-Blanco, Nereida
  Rodr{\'\i}guez-Fern{\'a}ndez, Francisco Cedr{\'o}n, Francisco~J Novoa, Adrian
  Carballal, Victor Maojo, Alejandro Pazos, and Carlos Fernandez-Lozano.
\newblock A review on machine learning approaches and trends in drug discovery.
\newblock {\em Computational and structural biotechnology journal},
  19:4538--4558, 2021.

\bibitem{shorten2021text}
Connor Shorten, Taghi~M Khoshgoftaar, and Borko Furht.
\newblock Text data augmentation for deep learning.
\newblock {\em Journal of big Data}, 8(1):1--34, 2021.

\bibitem{meyers2021novo}
Joshua Meyers, Benedek Fabian, and Nathan Brown.
\newblock De novo molecular design and generative models.
\newblock {\em Drug Discovery Today}, 26(11):2707--2715, 2021.

\bibitem{gupta2018generative}
Anvita Gupta, Alex~T M{\"u}ller, Berend~JH Huisman, Jens~A Fuchs, Petra
  Schneider, and Gisbert Schneider.
\newblock Generative recurrent networks for de novo drug design.
\newblock {\em Molecular informatics}, 37(1-2):1700111, 2018.

\bibitem{li2018multi}
Yibo Li, Liangren Zhang, and Zhenming Liu.
\newblock Multi-objective de novo drug design with conditional graph generative
  model.
\newblock {\em Journal of cheminformatics}, 10(1):1--24, 2018.

\bibitem{sousa2021generative}
Tiago Sousa, Jo{\~a}o Correia, V{\'\i}tor Pereira, and Miguel Rocha.
\newblock Generative deep learning for targeted compound design.
\newblock {\em Journal of Chemical Information and Modeling},
  61(11):5343--5361, 2021.

\bibitem{o2012towards}
Noel~M O’Boyle.
\newblock Towards a universal smiles representation-a standard method to
  generate canonical smiles based on the inchi.
\newblock {\em Journal of cheminformatics}, 4(1):1--14, 2012.

\bibitem{o2018deepsmiles}
Noel O'Boyle and Andrew Dalke.
\newblock Deepsmiles: an adaptation of smiles for use in machine-learning of
  chemical structures.
\newblock {\em chemrxiv.org}, 2018.

\bibitem{krenn2020self}
Mario Krenn, Florian H{\"a}se, AkshatKumar Nigam, Pascal Friederich, and Alan
  Aspuru-Guzik.
\newblock Self-referencing embedded strings (selfies): A 100\% robust molecular
  string representation.
\newblock {\em Machine Learning: Science and Technology}, 1(4):045024, 2020.

\bibitem{irwin2005zinc}
John~J Irwin and Brian~K Shoichet.
\newblock Zinc- a free database of commercially available compounds for virtual
  screening.
\newblock {\em Journal of chemical information and modeling}, 45(1):177--182,
  2005.

\bibitem{sterling2015zinc}
Teague Sterling and John~J Irwin.
\newblock Zinc 15--ligand discovery for everyone.
\newblock {\em Journal of chemical information and modeling},
  55(11):2324--2337, 2015.

\bibitem{jin2018junction}
Wengong Jin, Regina Barzilay, and Tommi Jaakkola.
\newblock Junction tree variational autoencoder for molecular graph generation.
\newblock In {\em International conference on machine learning}, pages
  2323--2332. PMLR, 2018.

\bibitem{wen2022fingerprints}
Naifeng Wen, Guanqun Liu, Jie Zhang, Rubo Zhang, Yating Fu, and Xu~Han.
\newblock A fingerprints based molecular property prediction method using the
  bert model.
\newblock {\em Journal of Cheminformatics}, 14(1):1--13, 2022.

\bibitem{segler2018rnn}
Marwin~HS Segler, Thierry Kogej, Christian Tyrchan, and Mark~P Waller.
\newblock Generating focused molecule libraries for drug discovery with
  recurrent neural networks.
\newblock {\em ACS central science}, 4(1):120--131, 2018.

\bibitem{fingerprint2015}
Adri{\`a} Cereto-Massagu{\'e}, Mar{\'\i}a~Jos{\'e} Ojeda, Cristina Valls,
  Miquel Mulero, Santiago Garcia-Vallv{\'e}, and Gerard Pujadas.
\newblock Molecular fingerprint similarity search in virtual screening.
\newblock {\em Methods}, 71:58--63, 2015.

\bibitem{ghose1999knowledge}
Arup~K Ghose, Vellarkad~N Viswanadhan, and John~J Wendoloski.
\newblock A knowledge-based approach in designing combinatorial or medicinal
  chemistry libraries for drug discovery. 1. a qualitative and quantitative
  characterization of known drug databases.
\newblock {\em Journal of combinatorial chemistry}, 1(1):55--68, 1999.

\bibitem{wildman1999prediction}
Scott~A Wildman and Gordon~M Crippen.
\newblock Prediction of physicochemical parameters by atomic contributions.
\newblock {\em Journal of chemical information and computer sciences},
  39(5):868--873, 1999.

\bibitem{lipinski2004lead}
Christopher~A Lipinski.
\newblock Lead-and drug-like compounds: the rule-of-five revolution.
\newblock {\em Drug discovery today: Technologies}, 1(4):337--341, 2004.

\bibitem{craig1998molecular}
David~Parker Craig and Thiru Thirunamachandran.
\newblock {\em Molecular quantum electrodynamics: an introduction to
  radiation-molecule interactions}.
\newblock Courier Corporation, 1998.

\bibitem{ertl2009estimation}
Peter Ertl and Ansgar Schuffenhauer.
\newblock Estimation of synthetic accessibility score of drug-like molecules
  based on molecular complexity and fragment contributions.
\newblock {\em Journal of cheminformatics}, 1(1):1--11, 2009.

\bibitem{2017rnn}
Esben~Jannik Bjerrum and Richard Threlfall.
\newblock Molecular generation with recurrent neural networks (rnns).
\newblock {\em arXiv preprint arXiv:1705.04612}, 2017.

\bibitem{BIMODAL}
Francesca Grisoni, Michael Moret, Robin Lingwood, and Gisbert Schneider.
\newblock Bidirectional molecule generation with recurrent neural networks.
\newblock {\em Journal of chemical information and modeling}, 60(3):1175--1183,
  2020.

\bibitem{reinvent2}
Thomas Blaschke, Josep Ar{\'u}s-Pous, Hongming Chen, Christian Margreitter,
  Christian Tyrchan, Ola Engkvist, Kostas Papadopoulos, and Atanas Patronov.
\newblock Reinvent 2.0: an ai tool for de novo drug design.
\newblock {\em Journal of chemical information and modeling},
  60(12):5918--5922, 2020.

\bibitem{ORGAN}
Gabriel~Lima Guimaraes, Benjamin Sanchez-Lengeling, Carlos Outeiral, Pedro
  Luis~Cunha Farias, and Al{\'a}n Aspuru-Guzik.
\newblock Objective-reinforced generative adversarial networks (organ) for
  sequence generation models.
\newblock {\em arXiv preprint arXiv:1705.10843}, 2017.

\bibitem{chembl}
Anna Gaulton, Louisa~J Bellis, A~Patricia Bento, Jon Chambers, Mark Davies,
  Anne Hersey, Yvonne Light, Shaun McGlinchey, David Michalovich, Bissan
  Al-Lazikani, et~al.
\newblock Chembl: a large-scale bioactivity database for drug discovery.
\newblock {\em Nucleic acids research}, 40(D1):D1100--D1107, 2012.

\bibitem{gomez2018automatic}
Rafael G{\'o}mez-Bombarelli, Jennifer~N Wei, David Duvenaud, Jos{\'e}~Miguel
  Hern{\'a}ndez-Lobato, Benjam{\'\i}n S{\'a}nchez-Lengeling, Dennis Sheberla,
  Jorge Aguilera-Iparraguirre, Timothy~D Hirzel, Ryan~P Adams, and Al{\'a}n
  Aspuru-Guzik.
\newblock Automatic chemical design using a data-driven continuous
  representation of molecules.
\newblock {\em ACS central science}, 4(2):268--276, 2018.

\bibitem{qm9}
Raghunathan Ramakrishnan, Pavlo~O Dral, Matthias Rupp, and O~Anatole
  Von~Lilienfeld.
\newblock Quantum chemistry structures and properties of 134 kilo molecules.
\newblock {\em Scientific data}, 1(1):1--7, 2014.

\bibitem{simonovsky2018graphvae}
Martin Simonovsky and Nikos Komodakis.
\newblock Graphvae: Towards generation of small graphs using variational
  autoencoders.
\newblock In {\em International conference on artificial neural networks},
  pages 412--422. Springer, 2018.

\bibitem{molecularrnn}
Mariya Popova, Mykhailo Shvets, Junier Oliva, and Olexandr Isayev.
\newblock Molecularrnn: Generating realistic molecular graphs with optimized
  properties.
\newblock {\em arXiv preprint arXiv:1905.13372}, 2019.

\bibitem{de2018molgan}
Nicola De~Cao and Thomas Kipf.
\newblock Molgan: An implicit generative model for small molecular graphs.
\newblock {\em arXiv preprint arXiv:1805.11973}, 2018.

\bibitem{gcpn}
Jiaxuan You, Bowen Liu, Zhitao Ying, Vijay Pande, and Jure Leskovec.
\newblock Graph convolutional policy network for goal-directed molecular graph
  generation.
\newblock {\em Advances in neural information processing systems}, 31, 2018.

\bibitem{landrum2013rdkit}
Greg Landrum et~al.
\newblock Rdkit: A software suite for cheminformatics, computational chemistry,
  and predictive modeling.
\newblock {\em Greg Landrum}, 2013.

\bibitem{doersch2016tutorial}
Carl Doersch.
\newblock Tutorial on variational autoencoders.
\newblock {\em arXiv preprint arXiv:1606.05908}, 2016.

\bibitem{pu2016variational}
Yunchen Pu, Zhe Gan, Ricardo Henao, Xin Yuan, Chunyuan Li, Andrew Stevens, and
  Lawrence Carin.
\newblock Variational autoencoder for deep learning of images, labels and
  captions.
\newblock {\em Advances in neural information processing systems}, 29, 2016.

\bibitem{burgess2018understanding}
Christopher~P Burgess, Irina Higgins, Arka Pal, Loic Matthey, Nick Watters,
  Guillaume Desjardins, and Alexander Lerchner.
\newblock Understanding disentangling in $\beta$-vae.
\newblock {\em arXiv preprint arXiv:1804.03599}, 2018.

\bibitem{mathieu2019disentangling}
Emile Mathieu, Tom Rainforth, Nana Siddharth, and Yee~Whye Teh.
\newblock Disentangling disentanglement in variational autoencoders.
\newblock In {\em International Conference on Machine Learning}, pages
  4402--4412. PMLR, 2019.

\bibitem{rydhmer2021dynamic}
Klas Rydhmer and Raghavendra Selvan.
\newblock Dynamic $\beta$-vaes for quantifying biodiversity by clustering
  optically recorded insect signals.
\newblock {\em Ecological Informatics}, 66:101456, 2021.

\bibitem{bowman2015generating}
Samuel~R Bowman, Luke Vilnis, Oriol Vinyals, Andrew~M Dai, Rafal Jozefowicz,
  and Samy Bengio.
\newblock Generating sentences from a continuous space.
\newblock {\em arXiv preprint arXiv:1511.06349}, 2015.

\bibitem{lim2018molecular}
Jaechang Lim, Seongok Ryu, Jin~Woo Kim, and Woo~Youn Kim.
\newblock Molecular generative model based on conditional variational
  autoencoder for de novo molecular design.
\newblock {\em Journal of cheminformatics}, 10(1):1--9, 2018.

\bibitem{gdtuo}
Kartik Chandra, Erik Meijer, Samantha Andow, Emilio Arroyo{-}Fang, Irene Dea,
  Johann George, Melissa Grueter, Basil Hosmer, Steffi Stumpos, Alanna Tempest,
  and Shannon Yang.
\newblock Gradient descent: The ultimate optimizer.
\newblock {\em CoRR}, abs/1909.13371, 2019.

\bibitem{samanta2020nevae}
Bidisha Samanta, Abir De, Gourhari Jana, Vicen{\c{c}} G{\'o}mez, Pratim~Kumar
  Chattaraj, Niloy Ganguly, and Manuel Gomez-Rodriguez.
\newblock Nevae: A deep generative model for molecular graphs.
\newblock {\em The Journal of Machine Learning Research}, 21(1):4556--4588,
  2020.

\bibitem{mapping}
Jos{\'e} Jim{\'e}nez-Luna, Francesca Grisoni, and Gisbert Schneider.
\newblock Drug discovery with explainable artificial intelligence.
\newblock {\em Nature Machine Intelligence}, 2(10):573--584, 2020.

\bibitem{shao2020controlvae}
Huajie Shao, Shuochao Yao, Dachun Sun, Aston Zhang, Shengzhong Liu, Dongxin
  Liu, Jun Wang, and Tarek Abdelzaher.
\newblock Controlvae: Controllable variational autoencoder.
\newblock In {\em International Conference on Machine Learning}, pages
  8655--8664. PMLR, 2020.

\bibitem{shao2021controlvae}
Huajie Shao, Zhisheng Xiao, Shuochao Yao, Dachun Sun, Aston Zhang, Shengzhong
  Liu, Tianshi Wang, Jinyang Li, and Tarek Abdelzaher.
\newblock Controlvae: Tuning, analytical properties, and performance analysis.
\newblock {\em IEEE transactions on pattern analysis and machine intelligence},
  2021.

\bibitem{ang2018proposed}
Guang Jun~Nicholas Ang, Wei~Ze Lim, and Choo~Min Lim.
\newblock A proposed two-input two-output self-tuning control scheme.
\newblock In {\em 2018 IEEE Symposium on Computer Applications \& Industrial
  Electronics (ISCAIE)}, pages 102--107. IEEE, 2018.

\end{thebibliography}
\end{document}